\documentclass{article}
\usepackage{iclr2023_conference,times}

\usepackage{amsmath,amsfonts,bm}

\def\eqref#1{equation~\ref{#1}}

\def\1{\bm{1}}

\DeclareMathAlphabet{\mathsfit}{\encodingdefault}{\sfdefault}{m}{sl}
\SetMathAlphabet{\mathsfit}{bold}{\encodingdefault}{\sfdefault}{bx}{n}

\usepackage{hyperref}
\usepackage{url}
\usepackage{enumitem}
\usepackage{booktabs}

\usepackage{graphicx}
\usepackage{caption}
\usepackage{subcaption}
\usepackage{multirow}
\usepackage{adjustbox}

\newcommand{\scaletables}{0.91}

\title{CodeGen2: Lessons for Training LLMs on Programming and Natural Languages}

\iclrfinalcopy

\author{Erik Nijkamp\thanks{\hspace{4pt}Equal contribution.},\enspace{\bf Hiroaki Hayashi}\footnotemark[1],\enspace{\bf Caiming Xiong},\enspace{\bf Silvio Savarese},\enspace{\bf Yingbo Zhou}\\
\\Salesforce Research}

\begin{document}

\maketitle

\begin{abstract}
Large language models (LLMs) have demonstrated remarkable abilities in representation learning for program synthesis and understanding tasks. The quality of the learned representations appears to be dictated by the neural scaling laws as a function of the number of model parameters and observations, while imposing upper bounds on the model performance by the amount of available data and compute, which is costly.

In this study, we attempt to render the training of LLMs for program synthesis more efficient by unifying four key components: (1) model architectures, (2) learning methods, (3) infill sampling, and, (4) data distributions. Specifically, for the model architecture, we attempt to unify encoder and decoder-based models into a single prefix-LM. For learning methods, (i) causal language modeling, (ii) span corruption, (iii) infilling are unified into a simple learning algorithm. For infill sampling, we explore the claim of a "free lunch" hypothesis. For data distributions, the effect of a mixture distribution and multi-epoch training of programming and natural languages on model performance is explored.

We conduct a comprehensive series of empirical experiments on 1B LLMs, for which failures and successes of this exploration are distilled into five lessons. We will provide a final recipe for training and release CodeGen2 models in size 1B, 3.7B, 7B, and, 16B parameters, along with the training framework as open-source: \url{https://github.com/salesforce/CodeGen}.
\end{abstract}

\section{Introduction}

\subsection{Motivation: Cost of LLMs}

Large language models (LLMs) have demonstrated strong empirical performance in a myriad of tasks across domains. In recent work~\citep{codex, nijkamp2022codegen, fried2022incoder, allal2023santacoder}, these findings have been transferred from natural to programming languages and achieved impressive performance in program synthesis and understanding tasks. The appeal of these models stems from three properties: (1) simple - the involved architectures are of low technical complexity due to reliance on self-attention circuits, (2) universal - that is a single model can handle a variety of different tasks rather than $n$ specialized models for $n$ tasks, which dramatically reduces resource and cost requirements, and, (3) scale - that is neural scaling laws dictate the performance of the model as a function of the amount of model parameters, data, compute in the form of power laws, that is, larger models usually yield predictably improved performance on down-stream tasks.

However, these merits obscure unresolved challenges: (1) while the self-attention circuit is technically simple, one has to pick an attention masking scheme to learn either bi-directional representations (encoder) or uni-directional representations (decoder), (2) while transformers appear task-agnostic, synthesis and understanding tasks have not been unified, (3) while improving performance with scale is attractive, even training a small set of models for different tasks incurs a significant cost. For the practitioner, the choices of model architecture, learning algorithm, and data distributions are not obvious. Exploration of these choices induces high monetary cost stemming from compute requirements.

\subsection{Goals: Reduce Cost by Unification and Open-source}

To address the concerns of monetary cost and choice of variants, we attempt to unify (1) model architecture, (2) learning objective, (3) left-to-right and infill sampling, and (4) data distributions into a single recipe, which yields a single universal model with competitive performance on a wide range of synthesis and understanding tasks.

To approach the unification of these aspects in a principled manner, we postulate and evaluate the following hypotheses:

\begin{enumerate}[label={(\arabic*)}]
    \item \textbf{Model Architecture:} Encoder and decoder representations can be unified into a Prefix-LM~\citep{raffel2020exploring} for which the bi-directional self-attention is beneficial for harder few-shot tasks without degradation in performance over standard causal-decoders.
    \item \textbf{Learning Algorithm:} A mixture of objectives for causal language modeling and span-corruption yields efficient information transport for zero-shot learning (decoder) and understanding tasks (encoder). 
    \item \textbf{Sampling Procedure:} Equipping a model with both left-to-right and infill sampling, under the assumption of the "free lunch" hypothesis, does not increase computational cost.
    \item \textbf{Data Distributions:} A mixture of natural and programming languages simultaneously benefits tasks in both domains without affecting performance within a single modality.
\end{enumerate}

The goals of this work are (i) to share lessons and provide a unified recipe to train such a universal model, (ii) to open-source implementation of the training procedure, and, (iii) to open-source a family of well-trained models.

\subsection{Findings: A Mixed Bag of Results}

In our attempt to achieve these goals, our method is to attempt full unification across all aspects and collect evidence to guide the ablation of features. We attempt to provide evidence to reject or not reject these hypotheses in an extensive study across a large set of experiments on 1B LLMs. The findings for our postulated hypotheses are summarized as follows:

\begin{enumerate}[label={(\arabic*)}]
    \item \textbf{Model Architecture:} We failed to provide evidence to quantify any benefits of Prefix-LM over the causal-decoder baseline within our set of evaluation tasks.
    \item \textbf{Learning Algorithm:} We succeeded in a simple mixture of objective functions while maintaining zero-shot performance.
    \item \textbf{Sampling Procedure:} We failed to provide evidence for the "free lunch" hypothesis of equipping models with infill sampling without incurring an additional cost in compute.
    \item \textbf{Data Distributions:} We show promising evidence of mixing natural and programming languages into a single model. We provide strong results for multi-epoch training.
\end{enumerate}

While we did not achieve full unification, we gained valuable findings and trained competitive open-source models on permissive data.

\subsection{Contributions: Lessons, Recipe, and Open-Source}

We share these findings distilled with the following contributions:

\begin{itemize}
    \item \textbf{Five lessons:} Distillation of results on (1) Prefix-LM as an architecture, (2) Free-lunch hypothesis of infill sampling, (3) Choice of objective functions, (4) Data mixture of natural and programming language, and, (5) Multi-epoch training,
    \item \textbf{Simple mixture objective:} We propose a simple, unified mixture of uncorrupted and within-file span-corruption sequences with next-token-prediction which yields competitive performance for both left-to-right and fill-in-the-middle auto-regressive sampling,
    \item \textbf{Open-Source implementation:} We will provide a well-engineered and tested reference implementation for LLM training of the final recipe,
    \item \textbf{Open-Source models:} We will open-source the CodeGen2 family of infill-capable models trained solely on permissive data once training for larger LLMs has converged.
\end{itemize}

\subsection{Related Work}

\paragraph{LLMs on Code} Transformers capture dependency among sequence elements through attention mechanism \citep{bahdanau2014neural} and are highly scalable, as shown in natural language processing \citep{devlin2018bert, lewis2020bart, raffel2020exploring}. Several efforts explore these models for program synthesis~\citep{chen2021evaluating, austin2021program, alphacode, fried2022incoder, nijkamp2022codegen, allal2023santacoder} and its effectiveness~\citep{vaithilingam2022expectation}.

\paragraph{Ablation Studies}

\cite{raffel2020exploring} introduce the concept of non-causal decoder in the form of a Prefix-LM with favorable performance over causal decoders after fine-tuning on down-stream tasks. The performance in few-shot generative tasks was not evaluated. \cite{wang2022language} conduct an extensive ablation study over architectures and objectives with the conclusion that decoder-only models with causal language modeling exhibit the strongest zero-shot generalization. Therefore, we limit our investigation to causal and non-causal decoders. \cite{tay2022unifying} compare encoder-decoder, decoder-only, and Prefix-LM architectures and report the beneficial performance of encoder-decoder models, while zero-shot generation tasks are not evaluated. The authors later adopt Prefix-LM instead of encoder-decoder in~\citep{tay2022transcending}.

\paragraph{Data Mixtures}

LaMDA~\citep{thoppilan2022lamda} was trained on a mixture of various data sources including dialogues, code documents, Q\&A data, tutorials, and, Wikipedia. However, the impact of this mixture and the specific sources are unclear. \cite{xie2023data} introduces a data selection method based on importance resampling which allows to mix datasets of various sizes, however, the evaluation only covers encoder-only models.

\section{Method: From Unification to Ablation}

In this Section, requirements for our goals are defined along with relevant components for ablation.

\subsection{Requirement: Performance on a Variety of Tasks}

We aim to render the training of LLMs for program synthesis more efficient by providing a unification of both learning methods and model architectures, while maintaining (or improving) performance of individual tasks under identical compute budget. The set of tasks is as follows:

\begin{enumerate}[label={(\arabic*)}]
    \item \textbf{Program Synthesis with left-to-right sampling (zero-shot)} The prompt as an intent specification is in the form of a function signature and doc-string. A program is conditionally sampled (or completed) based on the prompt in left-to-right auto-regressive fashion. The HumanEval~\citep{codex} is recruited to evaluate the quality of synthesized programs. Specifically, for prompt $(a)$, we sample $b \sim P(b|a)$.
    \item \textbf{Program Synthesis with infill sampling (zero-shot)} The prompt includes both past and future tokens for which the tokens "in-between" are supposed to be sampled. For instance, the prompt includes the first and last few lines of a function definition, while the body of the function is to be filled in. Specifically, for prompt $(a,c)$, we sample $b \sim P(b|a,c)$. The HumanEval-Infill~\citep{fried2022incoder} benchmark is recruited for evaluation.
    \item \textbf{In-context Learning from examples (few-shot)} A task is defined given a set of examples (or "shots"). Specially, for $n$ few-shot examples $((x_1, y_1), \ldots, (x_n, y_n))$ with code $x$ and label $y$, we sample label $y_{n+1} \sim P(y|(x_1, y_1), \ldots, (x_n, y_n), (x_{n+1}))$. The XSum benchmark~\citep{narayan2018don} is recruited for evaluation.
    \item \textbf{Program Understanding with bi-directional representations (fine-tune)} Causally-masked decoder models are constrained to auto-regressive sampling in left-to-right fashion. For understanding tasks, such as defect detection~\citep{lu2021codexglue}, removing this constraint over time such that the representations can be a function of all input tokens simultaneously seems desirable. Such representations are obtained from bi-directional language models without causal masking. The CodeXGLUE~\citep{lu2021codexglue} and SuperGLUE~\citep{wang2019superglue} benchmarks are recruited for evaluation.  
\end{enumerate}

\subsection{Components: Architecture, Objective, Sampling, Data}

\paragraph{Model Architecture}

In representation learning with transformers~\citep{devlin2018bert, lewis2020bart, raffel2020exploring}, two schemes of modeling are prevalent which differ in their attention masks for the contextualization of hidden vectors. For a sequence $x=(x_1, \ldots, x_n)$ of $n$ vectors, we differ: (1) bi-directional encoder-based representations in which each token vector $x_i$ can attend all other tokens $\{x_j : i=1,\ldots, n\}$, (2) uni-directional decoder-based representations in which each token vector $x_i$ can only attend previous tokens $ \{x_j : j \leq i\}$. While encoder-based representations for which each hidden vector can contextualize with all other vectors may be desirable for understanding tasks, decoder-based representations with temporal causal masking are required for language modeling for which the joint density is factorized as the product of conditionals over time steps. To unify both schemes, we adopt the notion of prefix-based language modeling (Prefix-LM)~\citep{raffel2020exploring}. For a prefix, we decompose the input sequence $x$ into a prefix $p$ and a context $c$. For the prefix $p=(x_1, \ldots, x_m)$ where $m<n$, each token can attend over all other tokens in the prefix, which amounts to bi-directional representations. For the context $c=(x_{m+1}, \ldots, x_n)$, each token can only attend to previous tokens, which amounts to uni-directional decoder representations. This unifies bi-directional attention over the prefix with the requirement of causal masking to factorize the joint density over time. The hope is to achieve competitive auto-regressive sampling for synthesis tasks, while learning strong bi-directional representations for understanding tasks.

\paragraph{Learning Algorithm}

The choice of encoder or decoder-based model architectures typically guides the selection of learning algorithms for language modeling. Encoder-based models may be trained with the task of masked language modeling in the form of denoising span corruptions~\citep{devlin2018bert,raffel2020exploring}. Decoder-based models may be trained as density language modeling in the form of a next-token-prediction task~\citep{radford2018improving}. For encoder-based models, the prevalent algorithms are variants of token reconstructions or denoising task for which spans of tokens undergo a corruption or perturbation. For a sequence $x=(x_1, \ldots, x_n)$ of $n$ tokens, a perturbation $\tilde{x}=(x_1, m_1, x_5, x_6, m_2, x_7, \ldots, x_n)$ replaces spans of tokens with special mask tokens $(m_1, m_2, \ldots)$. The learning task is to recover the original sequence $x$ from the perturbation $\tilde{x}$. Denoising-based learning objectives have been shown to be highly efficient for language understanding tasks. For decoder-based models, the prevalent algorithm is maximum likelihood-based learning of causal language modeling in the form of a next token prediction task~\citep{radford2018improving}. For a sequence $x=(x_1, \ldots, x_n)$ of $n$ tokens, the task is to predict the token $x_i$ given previous tokens $(x_j: j < i)$. 
In this work, we explore a learning algorithm as a mixture of both causal language modeling objectives and span corruption. We postulate for such a mixture the task-specific prior information should be minimized to avoid over-fitting to specific tasks. That is, ideally, the distributions over mixture ratio of tasks, length of the prefixes, and length of spans are uniform.

\paragraph{Sampling Procedure} Program synthesis in the form of auto-regressive sampling from a language model has been established as a predominant method. While left-to-right sampling can only take previous tokens into account, often when editing existing files of code, conditioning the sampling on context before and after the current position within a file is desirable. Several variants~\citep{du2022glm, fried2022incoder, bavarian2022efficient} which rearrange a sequence $(a,b,c)$ into $(a,{<}mask{>},c,{<}end{>},b)$ such that the infilling $b$ given $(a,c)$ can be learned by standard next-token-prediction objectives in causally masked decoders. \cite{bavarian2022efficient} claim under the "free lunch" hypothesis that training LLMs with such modified training observation does not incur any additional cost in compute or degradation in performance on zero-shot generation tasks.

\paragraph{Data Distribution} In maximum likelihood learning, a model is learned by minimizing some measure of divergence between the data and model distribution. Surprisingly, when increasing the number of observations and model parameters, few-shot abilities emerge when sampling from the fitted densities~\citep{brown2020language, wei2022emergent}. For program synthesis, \cite{nijkamp2022codegen} demonstrate sampling executable code in a multi-turn conversational scheme, similar to \citep{ouyang2022training}, but without explicit instruction fine-tuning. It is hypothesized that these abilities emerge from weak supervision in the data. Programs often include functions with accompanying English descriptions. We attempt to increase the amount of natural language to further improve this ability and explicitly create a mixture of natural and programming languages. Further, the resulting model may be competitive on down-stream tasks in both domains.

\section{Results: Lessons and Recipe}

In this Section, we present the empirical results and conclusions of the attempt toward unification and ablation study for which the findings are distilled into the following five lessons.

\subsection{Lesson 1: Prefix-LM's benefit is questionable}

As discussed, Prefix-LM behaves as an encoder with bi-directional attention, and as an autoregressive decoder with a causal mask.
However, it is unclear if this unification of the architecture leads to competitive performance on both ends.
We evaluate this question from three angles: data, representation, and objective.
In the following, we refer to the first half of the sequence that is covered by bi-directional attention as \textit{non-causal} part, and the rest of the sequence as \textit{causal} part.

\subsubsection{Data}

\paragraph{Context and Hypothesis}
When training a Prefix-LM with next token prediction, the loss for the non-causal part is masked because the prediction is informed by the future tokens that are used when encoding.
Depending on the length of the non-causal part in each sequence, this means the reduction of the effective number of tokens responsible for the gradient update, which raises the question of whether or not the prefix mask negatively impacts the learning of Prefix-LM. Specifically, we hypothesize that the lack of gradient on the non-causal part results in worse results on the code generation task due to the slower rate of information transport via NTP.

\paragraph{Results and Findings}
To evaluate this hypothesis, we first train a Prefix-LM on BigPython~\citep{nijkamp2022codegen} with causal language modeling and compare the model performance on HumanEval against a causal decoder baseline trained for the same number of steps.
The length of the non-causal part is set to $rN$, where $r \sim [0, 0.9]$ and $N$ is the sequence length.
Surprisingly, except for the first 100,000 steps, we observe that the HumanEval scores for Prefix-LM are on-par with the causal decoder for the majority of pre-training, not exhibiting a tendency of slower learning.
In addition, Prefix-LM trained with the combination of causal language modeling and span corruption yields competitive infill capability, relative to InCoder~\citep{fried2022incoder}.

Next, we repeat the same experiment on the Stack~\citep{kocetkov2022stack}, with Python being 9.4\% of all tokens in the dataset.
We refer to Table~\ref{tab:Prefix-LM_comparison} for HumanEval and HumanEval-Infill results.
Contrary to the promising result above, however, we observed strictly worse pass@$k$ throughout the pre-training, resulting in 2 points worse than the causal decoder baseline. 

Based on these conflicting observations, we speculate that a negative effect due to the lack of enough gradient update for the target language (\textit{e.g.} Python) exists in Prefix-LM, which did not surface earlier thanks to training exclusively on Python for a large number of steps.
It still remains a question as to whether this effect surfaces at a certain threshold or gradually.

\subsubsection{Representation}

\paragraph{Context and Hypothesis}
Prefix-LM is appealing due to the bi-directional attention, which allows for the model to contextualize hidden states with both past and future tokens.
Controlled by only the existence of an attention mask, the ease of switching the role as an encoder or decoder allows for trivial unification, if the resulting model is competitive.

Based on the recent success with Prefix-LM~\citep{tay2022transcending}, we hypothesize that the trained Prefix-LM yields a competitive encoder representation that can be used for a range of discriminative tasks, while retaining generative capability as the decoder.

To test this hypothesis, we train Prefix-LM with a mixture of causal language modeling and span corruption, given that denoising yields strong representations~\citep{raffel2020exploring,tay2022unifying}.
Aiming for wider coverage of tasks, we train the model on both programming and natural language and evaluate each model on the defect detection task from CodeXGLUE~\citep{lu2021codexglue} and 4 tasks from SuperGLUE~\citep{wang2019superglue}, respectively. 
We attach a classification head on top of the hidden state at a certain time step (first or last token) to adapt to each task.
In addition to the finetuning-based discriminative tasks, we examine if the in-context learning ability is enhanced by bi-directional attention.
We follow UL2 and evaluate the models trained on natural language on XSum as a few-shot generation task.
We examine the change in performance by increasing the number of examples fed into the model as context.
\paragraph{Results and Findings}
In Table~\ref{tab:Prefix-LM_comparison}, Prefix-LM generally achieves better results over causal decoder trained with the same objective on our SuperGLUE tasks, which partially answers our question that bi-directional attention may yield informative representations.
However, our 1B parameter model could not compete with much smaller encoder-only pre-trained models such as CodeBERT on CodeXGLUE or RoBERTa-large on SuperGLUE, which leads us to conclude that the representations are not sufficiently informative to justify the substitution of smaller scale encoder-only pre-trained models with one Prefix-LM.
For few-shot XSum, we did not observe meaningful differences between the two models, regardless of the number of exemplars in the non-causal part.

\begin{table}[tb]
\begin{center}
\scalebox{\scaletables}{
\begin{footnotesize}
\begin{tabular}{lcccc}
\toprule
\multirow{2}{*}{Model} & \multirow{2}{*}{Size} & \multirow{2}{*}{HumanEval} & \multicolumn{2}{c}{HumanEval-Infill} \\ \cmidrule{4-5}
                       &      &                     & SingleLine       & MultiLine \\\midrule
\multirow{2}{*}{Incoder}& 1.3B & \phantom{0}9.79           & 52.56            & 23.85     \\
                       & 6.7B & 15.20                      & 66.69            & 34.62     \\\midrule
\multirow{4}{*}{CodeGen2.0}& 1B   & 10.27                    & 53.41            & 23.41     \\
                       & 3.7B & 14.88                      & 65.72            & 33.45     \\
                       & 7B   & 19.09                      & 68.74            & 38.88     \\ 
                       & 16B   & 20.46                      & -            & -     \\\bottomrule
\end{tabular}
\end{footnotesize}
}
\end{center}
\caption{Pass@1 on HumanEval. For HumanEval-Infill, only end-of-mask truncation is used.}
\end{table}

\subsubsection{Objective}
\paragraph{Context and Hypothesis}
As discussed, Prefix-LM appears to be effective when trained with UL2~\citep{tay2022unifying}, which proposes to combine causal language modeling (CLM) and denoising with 6 different hyperparameters.
The objective achieves significant gain over CLM and T5-style span corruption on a range of tasks under different model scales.
In the hope of achieving similarly competitive results in the programming language domain, we train Prefix-LM with the UL2 objective (including the task tokens), but find that the trained model is significantly worse on HumanEval by 9 points on pass@1 compared to the CLM baseline.
We hypothesize that this unexpectedly under-performing result is due to the small ratio of non-corrupted sequences used by one of the UL2 objectives, S-denoiser, which on average treats 75\% of a sequence as the prefix.
\paragraph{Results and Findings}
We verify the effect of the prefix length by simplifying UL2's denoiser hyperparameters such that (1) the percentage of S-denoiser is higher (14\% to 50\%) and (2) the average prefix length for the S-denoiser is shorter (75\% to 50\%), and observe consistent improvement on HumanEval over the original UL2 hyperparameters.
In general, we conclude that causal language modeling on non-corrupted sequences is aligned with the zero-shot generation task in HumanEval, which requires sampling at the end of prefix, and therefore we maximize the number of tokens in the sequence used for gradient updates to learn a competitive model.

\subsection{Lesson 2: Infill is not a Free Lunch}
\paragraph{Context and Hypothesis}
The process of writing code involves editing tokens within the file, not only at the end. Among several works that have attempted to equip the model with this infill ability, \cite{bavarian2022efficient} claims to achieve infill with no degradation in left-to-right sampling at the end of context by carefully permuting a portion of sequences from training data, getting the infill ability ``for free.''
On the other hand, concurrent work reports that infill is ``cheap'', \textit{i.e.} there is a consistent performance drop in left-to-right sampling in HumanEval, if the model is trained on the infilling objective~\citep{allal2023santacoder}. 
While conflicting results have been reported, we hypothesize that indeed infill is for free, if trained following the methods by~\cite{bavarian2022efficient}, and attempt to reproduce infill learning with our models.
\paragraph{Results and Findings}
To verify the free-lunch hypothesis, we train a causal decoder with a mixture of CLM and PSM (Prefix, Suffix, Middle sequence reordering) infilling objective, following the experimental setting in~\cite{bavarian2022efficient}.
However, we observed about 1 point decrease in pass@1 in HumanEval performance compared to the causal decoder baseline trained only with causal language modeling.
Thus, our observation follows that of \citep{allal2023santacoder} in that infill is not for free, and we leave careful reimplementation to reproduce ``infill for free'' a future work.

\begin{table}[tb]
\begin{center}
\scalebox{\scaletables}{
\begin{footnotesize}
\begin{tabular}{lrrrrrr}
\toprule
\multirow{2}{*}{Model}         &  HumanEval & Infill-Single &  Infill-Multi &  XSum &  SuperGLUE & CodeXGLUE \\ 
         & pass@1 & pass@1          & pass@1      &  R-L  & BoolQ (Acc)    & Defect (Acc)  \\\midrule
Decoder  & 7.99   & 43.37           & 17.40       & 10.28       & 0.774          & 0.635        \\
Prefix-LM & 6.71   & 39.11           & 17.25       & 9.78        & 0.806          & 0.640        \\ \bottomrule
\end{tabular}
\end{footnotesize}
}
\end{center}
\caption{Comparison of causal decoder and Prefix-LM on various tasks. HumanEval, Infill-Single, and Infill-Multi are evaluated as zero-shot code generation, XSum is evaluated as 1-shot summarization, SuperGLUE (BoolQ) and CodeXGLUE (defect detection) are evaluated by fine-tuning.}
\label{tab:Prefix-LM_comparison}
\end{table}

\subsection{Lesson 3: Objective can be Simple, yet Needs to be Carefully Chosen}

\paragraph{Context and Hypothesis}
So far, we learn from the lessons above that (1) Prefix-LM is not an ideal architecture due to poor multilingual performance and sub-par utility of representations, (2) infill is not for free, but HumanEval performance is close to that of a non-infill model.

To train our causal decoder model with competitive left-to-right and infill sampling capability, we align the sequence format (e.g., no task tokens), remove complexity, and assume uniform distributions over task mixture and span lengths to avoid bias relative to \citep{tay2022unifying}. 
We choose span corruption as the base infill objective following InCoder~\citep{fried2022incoder}.
However, we take a different approach in selecting the spans for corruption: (1) we first sample a dynamic ratio of sequence to mask out, (2) we then sample the span length and mask out locations such that the total number of tokens match the ratio of the original sequence determined earlier.
Additionally, we introduce two changes below:
\begin{itemize}[leftmargin=*]
  \item \textbf{Mixed Objective.} While InCoder learns to predict the next tokens within subsequences appearing in the span-corrupted sequence, we incorporate a dedicated causal language modeling objective for a portion of sequences. For each sample, the choice of span corruption or causal language modeling objective is decided with $p=0.5$.
We do not prefix the sequence with a task token as proposed by UL2, since no notable differences are observed with the presence of task tokens.
  \item \textbf{File-level Corruption.}
Span corruption may accidentally mask the document boundaries, which leads to a malformed context that confuses the model.
To avoid this, we apply span corruption at file-level, \textit{i.e.}, if a sequence can be split into two subsequences because of a document boundary, apply the span corruption for each subsequence, and concatenate them together.
This results in some sequences having multiple pairs of denoising instances.
\end{itemize}
\paragraph{Results and Findings}

We examine our recipe on four model sizes: 1B, 3.7B, 7B, and 16B, and refer to them as \textbf{CodeGen2}.\footnote{As of submission, the 16B-parameter model is still under training. We will revise the manuscript and open-source the model once the training converges.}
For training a subset of the Stack v1.1~\citep{kocetkov2022stack}, filtered with a stronger permissive license guideline, is used.
The models are evaluated on HumanEval~\citep{chen2021evaluating} and HumanEval-Infill~\citep{fried2022incoder,bavarian2022efficient}, where we follow their truncation strategies.
However, we note the discrepancy in experimental results for InCoder from the original paper, as we only use the end-of-mask token (\texttt{<eom>}) instead of heuristics to determine the end of infill sequence to measure the line-agnostic ability to infill and stop.

\begin{figure*}[t]
    \centering
    \begin{adjustbox}{minipage=\linewidth,scale=0.7}
        \begin{subfigure}[b]{0.47\textwidth}
            \centering
            \includegraphics[width=0.98\textwidth]{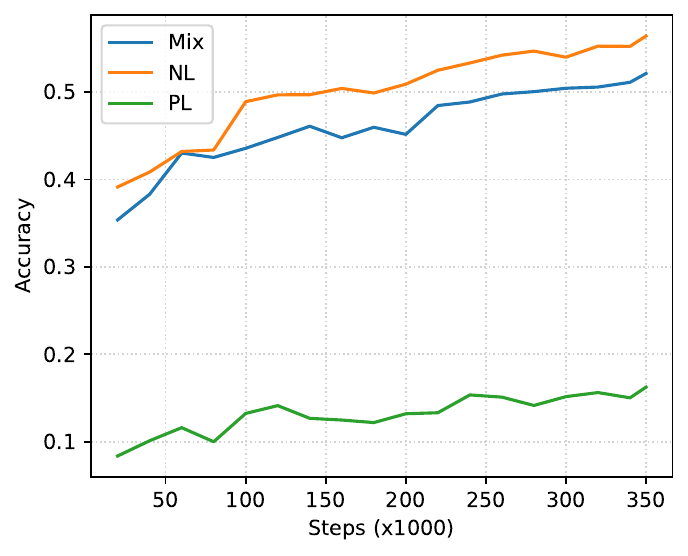}
            \caption[]%
            {{\small LAMBADA (Accuracy)}}    
            \label{fig:lambada_acc}
        \end{subfigure}
        \hfill
        \begin{subfigure}[b]{0.47\textwidth}  
            \centering 
            \includegraphics[width=\textwidth]{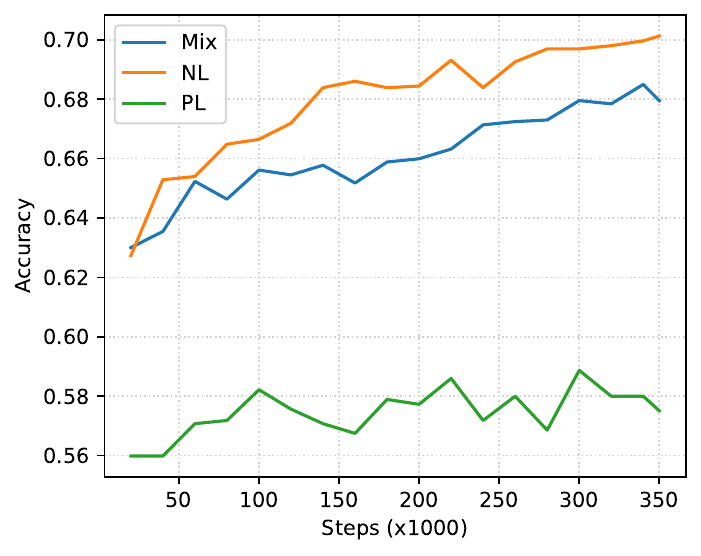}
            \caption[]%
            {{\small PIQA (Accuracy)}}    
            \label{fig:piqa_acc}
        \end{subfigure}
        \begin{subfigure}[b]{0.47\textwidth}   
            \centering 
            \includegraphics[width=0.97\textwidth]{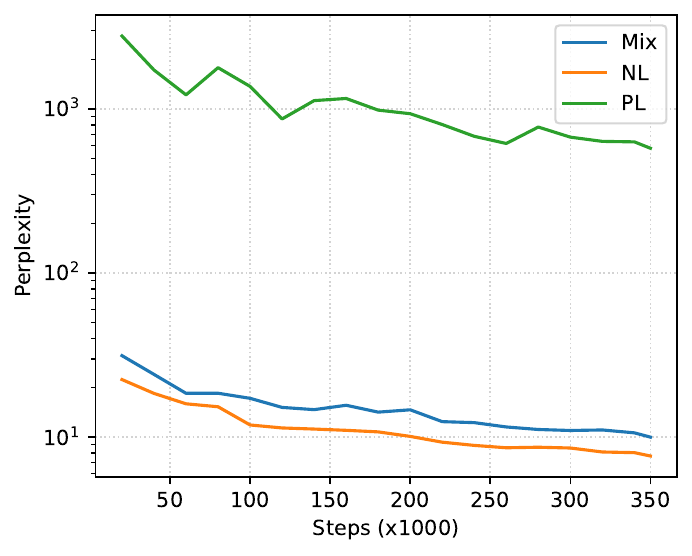}
            \caption[]%
            {{\small LAMBADA (Perplexity)}}    
            \label{fig:lambada_ppl}
        \end{subfigure}
        \hfill
        \begin{subfigure}[b]{0.47\textwidth}   
            \centering 
            \includegraphics[width=\textwidth]{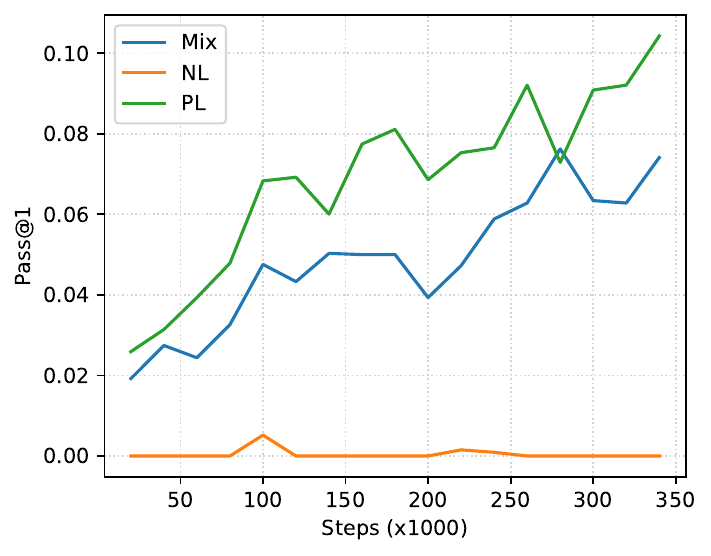}
            \caption[]%
            {{\small HumanEval (Pass@1)}}    
            \label{fig:he}
        \end{subfigure}
    \end{adjustbox}
    \caption[]
    {\small Results on LAMBADA and PIQA for NL and HumanEval for PL over number of training steps.} 
    \label{fig:mixing}
\end{figure*}

\subsection{Lesson 4: Multi-modal Data Mixing}
\paragraph{Context and Hypothesis}
With the growing interest in cross-modal applications between natural and programming languages due to the rise of code-aware conversational LLMs, there is a demand for tested training recipes on mixed textual modalities.
To some extent, an implicit mixture of modalities exists in programming language data in the form of comments and documents, and in natural language data as snippets of code for online QA forums, which is a sufficient supervised learning signal for a large-scale LLM for code to exhibit an ability to generate code from text~\citep{nijkamp2022codegen}. 
However, our preliminary experiment shows that such LLMs for code exhibit poor performance on natural language tasks, despite the implicit mixture.
It is our hypothesis that the poor performance can be mitigated with a sufficient learning signal from a distribution of natural language.
We test the hypothesis by training a causal decoder with the combined causal language modeling and span corruption objective on a mixture of natural and programming languages (\textbf{Mix}).
Examples in a batch are sampled from the Pile~\citep{gao2020pile} and the Stack data equally likely.

\paragraph{Results and Findings}
The model is evaluated in a zero-shot manner on both programming and natural language downstream tasks over the training steps to assess the effect of data mixing on each modality; specifically, we report pass@1 for HumanEval, accuracy and perplexity for LAMBADA~\citep{paperno-etal-2016-lambada}, and accuracy for PIQA~\citep{Bisk2020}.
For comparison, we alter the training data with (1) only the Pile~\citep{gao2020pile} (\textbf{NL}) or (2) only the Stack~\citep{kocetkov2022stack} (\textbf{PL}), and trained for the same number of steps. Fig~\ref{fig:mixing} depicts the results for these tasks. Consistently over the training steps, we observe that:
\begin{itemize}[leftmargin=*]
    \item \textbf{Mix} does not outperform other baselines for the evaluated tasks given the same compute budget.
    Since the exposure to each domain reduces by 50\% under \textbf{Mix}, the model reasonably performs worse than the domain-matched counterparts (\textit{e.g.}, \textbf{PL} for HumanEval).
    \item \textbf{Mix} performs closely to the domain-matched models. From early in the training, \textbf{Mix} substantially improves performance over the domain-mismatched baselines (\textit{e.g.}, \textbf{PL} for LAMBADA), which suggests that one should mix natural and programming languages if (a) the compute budget is constrained and (b) the resulting model is used for both domains.
\end{itemize}
Nonetheless, we find the recipe promising that even a simple mixture of training data allows for efficient learning on both domains, which might suggest that a slightly longer training with \textbf{Mix} can yield one competitive model for both domains.

\subsection{Lesson 5: Multi-epoch Training}
\paragraph{Context and Hypothesis}
Since the data in the wild is finite, the increase of scale in model size is bound at one point.
This is due to the common practice of pre-training the model for only one epoch, \textit{i.e.}, the model observes each sequence only once during training.
However, we hypothesize that the model has the capacity to absorb information from repeated observations. Moreover, span corruption can be interpreted as a means of data augmentation. That is, perturbations of the original observation may transport information under repetition. Finally, the extended learning rate decay function under an increased token budget allows for learning under higher learning rates over a longer span of time.

To test the hypothesis, we train a 7B parameter model with CodeGen2 objective on repeated StarCoderData~\citep{li2023starcoder}, a multilingual programming language data containing commits and issues, for a total of 1.4 trillion tokens, which amounts to 5 epochs.
The learning rate schedule is accordingly set so that the decay is over the total steps.
We refer to this model \textbf{CodeGen2.5}.

\paragraph{Results and Findings} 
Results are shown in Table~\ref{tab:cg25}.
Compared to the baseline CodeGen2-7B model that is trained for 400B tokens, we observe a significant gain in terms of pass@$k$, which suggest that multi-epoch training continues to improve the program synthesis capability despite not being exposed to new data after 280B.
However, it remains to be answered how each of the following factors contributes to the improvement: (1) more number of tokens, (2) span corruption as a form of data augmentation, (3) higher learning rate for a longer period.
To gain a more comprehensive understanding, it is crucial to conduct additional ablations, which we leave as future work.

\begin{table}[tb]
\begin{center}
\scalebox{\scaletables}{
\begin{footnotesize}
\begin{tabular}{lccccc}
\toprule
\multirow{2}{*}{Model} & \multirow{2}{*}{Size} & \multirow{2}{*}{Tokens} & \multicolumn{3}{c}{HumanEval}  \\ \cmidrule{4-6}
                       &      &     &  pass@1  & pass@10 & pass@100 \\\midrule
             {CodeGen2.0} & 7B   & 400B & 18.83   & 31.78   &  50.41 \\
             {CodeGen2.0} & 16B  & 400B & 20.46   & 36.50   &  56.71 \\
             {CodeGen2.5} & 7B   & 1.4T & 28.36   & 47.46   &  75.15 \\ \bottomrule
\end{tabular}
\end{footnotesize}
}
\end{center}
\caption{\label{tab:cg25}Pass@$k$ on HumanEval under temperatures $t=0.2, 0.6, 0.8$ for $k=1,10,100$, respectively.}
\end{table}

\section{Conclusion: Lessons and Open-Source}

The training of LLMs is costly and involves a myriad of design choices. Our goal was to address this challenge by unification across architecture, objectives, sampling procedures, and data distributions. We formed hypotheses for each of these aspects and distilled the positive and negative findings into five lessons. While we did not achieve satisfactory unification, the findings of this exploration and our final training recipe may be valuable for practitioners.

In summary, for our hypotheses, we believe (1) the Prefix-LM architecture does not yield any measurable improvements on our set of tasks, (2) training a model with infill sampling is not a free lunch, (3) a simple mixture of causal language modeling and span-corruption limited to within-file spans is sufficient, and (4) a mixture distribution of programming and natural languages looks promising, and, (5) there is strong evidence for the effectiveness of multi-epoch training as demonstrated with CodeGen2.5.

To facilitate future research, we will open-source the training implementation and the resulting family of CodeGen2 models with the sizes of 1B, 3.7B, 7B, and 16B parameters for the community.

\bibliography{iclr2023_conference}
\bibliographystyle{iclr2023_conference}

\end{document}